\documentclass[a4paper,UKenglish,cleveref, autoref]{lipics-v2019}

\title{A Phase Transition in Minesweeper} 

\titlerunning{Minesweeper Phase Transition}

\author{Ross Dempsey}{Joseph Henry Laboratories, Princeton University }{sdempsey@princeton.edu}{https://orcid.org/0000-0002-0881-8814}{}

\author{Charles Guinn}{Joseph Henry Laboratories, Princeton University }{cguinn@princeton.edu}{https://orcid.org/0000-0002-5337-5821}{}

\authorrunning{R. Dempsey \and C. Guinn}

\Copyright{Ross Dempsey and Charlie Guinn}



\category{}

\relatedversion{}

\supplement{}

\acknowledgements{The authors would like to thank Dumitru Caligaru and Katherine Xiang for interesting conversations and correspondence related to this project. We are also grateful to Anton Belov and Joao Marques-Silva for their open source GMUS extractor \texttt{MUSer2}.}

\nolinenumbers 

\hideLIPIcs  

\EventEditors{Martin Farach-Colton, Giuseppe Prencipe, and Ryuhei Uehara}
\EventNoEds{3}
\EventLongTitle{10th International Conference on Fun with Algorithms (FUN 2020)}
\EventShortTitle{FUN 2020}
\EventAcronym{FUN}
\EventYear{2020}
\EventDate{September 28--30, 2020}
\EventLocation{Favignana Island, Sicily, Italy}
\EventLogo{}
\SeriesVolume{157}
\ArticleNo{12}

\def\ms{\emph{Minesweeper}}
\def\sat{\texttt{SAT}}
\def\unsat{\texttt{UNSAT}}
\def\cons{\texttt{CONSISTENCY}}
\def\infer{\texttt{INFERENCE}}
\renewcommand{\d}[2]{\frac{d#1}{d#2}}

\usepackage{tikz}

\begin{document}

\maketitle

\begin{abstract}
We study the average-case complexity of the classic \ms\ game in which players deduce the locations of mines on a two-dimensional lattice. Playing \ms\ is known to be co-NP-complete. We show empirically that \ms\ exhibits a phase transition analogous to the well-studied \sat\ phase transition. Above the critical mine density it becomes almost impossible to play \ms\ by logical inference. We use a reduction to Boolean unsatisfiability to characterize the hardness of \ms\ instances, and show that the hardness peaks at the phase transition. Furthermore, we demonstrate algorithmic barriers at the phase transition for polynomial-time approaches to \ms\ inference. Finally, we comment on expectations for the asymptotic behavior of the phase transition.
\end{abstract}

\section{Introduction}

\ms\ is a single-player game originally released by Microsoft as part of the Windows 3.1 operating system. Players are presented with a rectangular lattice of covered squares, and behind some of these squares lie mines. The object of the game is to uncover all empty squares while never uncovering a mine. Players deduce the locations of mines from numbers on the empty squares which indicate how many mines lie adjacent.


\ms\ is famous as a mode of mindless procrastination. On the television show \emph{The Office}, Jim Halpert quips ``those mines aren't going to sweep themselves'' in reference to his boredom. This popular sentiment would seem to be at odds with complexity-theoretic studies of \ms. In Section \ref{sec:complexity} we review earlier results which show that playing \ms\ by logical inference is co-NP-complete \cite{Scott2011MinesweeperMN}, and deciding whether a \ms\ board is consistent with some assignment of mine squares is NP-complete \cite{Kaye2000MinesweeperIN}, or even Turing-complete for an infinite board size \cite{Kaye2000MinesweeperTuring}.

This discrepancy is remedied by classic results on phase transitions in NP-complete problems. While NP-complete problems are widely believed to be exponentially hard in the worst case, they can be much easier in typical cases. As a representative example, for random $k$-\sat\ with $n$ literals and $c$ clauses, it is found empirically that the probability $p(n, c)$ of a formula being satisfiable drops sharply to zero when the ratio $\ell = c/n$ reaches a certain threshold \cite{Gent1994TheSP}. In fact, Friedgut's theorem \cite{friedgut1999sharp, friedgut2005hunting} implies the existence of a function $\ell_*(n)$ such that for any $\epsilon>0$, as $n\to\infty$
\begin{equation}
    p\left(n, (1-\epsilon)\ell_*(n)n\right) \to 1,\qquad p\left(n, (1+\epsilon)\ell_*(n)\right)\to 0.
\end{equation}
It is suspected that $\ell_*(n)$ converges to a limiting critical value.

This phase transition is closely associated with the average-case complexity of \sat, or NP-complete problems more generally. Backtracking-based search algorithms for solving \sat\ require the largest number of branches for $\ell$ near $\ell_*$ \cite{Gent1994TheSP}, and a similar increase in hardness at the phase transition is observed in other constraint satisfaction problems \cite{Achlioptas_2008}. As long as $\ell$ is sufficiently separated from $\ell_*$, \sat\ can be solved relatively quickly in average cases.

In this paper, we present evidence for a similar phase transition in \ms\ by showing empirically that the number of mines which can be flagged by an inference-based solver rapidly declines near a critical mine density. Furthermore, in Section \ref{sec:transition} we define a metric for the hardness of \ms\ instances and show that this metric peaks at the transition point. In Section \ref{sec:algorithms}, we discuss a simple family of polynomial-time algorithms and show that the hardness metric we employ represents a natural barrier for these algorithms. Finally, in Section \ref{sec:asymptotic}, we comment on expectations for the asymptotic behavior of the phase transition.

\section{Worst-Case Complexity: \cons\ and \infer}\label{sec:complexity}

We begin with some definitions. Throughout this paper, when we refer to \emph{Minesweeper} we mean the classic Windows game instantiated on an $N\times N$ square lattice. The rules of the game are as follows.
\begin{definition}
A \ms\ board with size $N$ and mine density $\rho$ is an $N\times N$ array with $\lfloor N^2\rho\rfloor$ sites occupied by mines, and the remainder of sites empty or ``safe.'' A \ms\ instance is a \ms\ board with a subset of its sites uncovered. All the uncovered sites are labeled by the number of neighboring sites (including diagonal neighbors, and including wrap-around neighbors for sites on the edge) which contain mines. A move consists of one of the following:
\begin{itemize}
    \item Revealing a covered site. This results in loss of the game if the site contains a mine. Otherwise, the site is uncovered and its number of neighboring mines is revealed. If this number is zero, all neighboring sites are revealed for the player automatically.
    \item Flagging a covered site. This is a bookkeeping tool for sites which are known to contain mines.
\end{itemize}
The game is won when all empty sites are uncovered.
\end{definition}
When attempting to make inferences about the covered squares, it is helpful to adjust the labels of uncovered squares by subtracting the mines already accounted for with flags. We thus define the following \cite{becerra2015algorithmic}:
\begin{definition}
The effective label of a \ms\ site its is label minus its number of flagged neighbors.
\end{definition}

Two decision problems related to \ms\ have been investigated in the past. The first, historically speaking, is \cons\ \cite{Kaye2000MinesweeperIN}. The \cons\ problem asks whether a given \ms\ instance could be realized by some assignment of mines to the covered squares. Figure \ref{fig:inconsistent} shows an example of an inconsistent instance which could never arise in an actual game of \ms. 

Clearly \cons\ belongs to NP by reduction to \sat. We represent each covered site by a Boolean variable which is true if and only if the site contains a mine. Each site with at least one covered neighbor defines a Boolean constraint. The instance is consistent if and only if the conjunction of all these constraints forms a satisfiable formula \cite{Kaye2000MinesweeperIN}.

\begin{figure}
    \centering
    \begin{subfigure}[t]{0.48\linewidth}
        \centering
        \begin{tikzpicture}[scale=0.5]
            \begin{scope}[shift={(-.5,-.5)}]
                \fill[gray!30] (0,2) rectangle (3,3);
                \draw[thick] (-1,0) grid (4,5);
            \end{scope}
            \node at (0,1) {2};
            \node at (1,1) {3};
            \node at (2,1) {2};
            \draw[very thick,shift={(0,2)},fill=red] (.1,-0.3) -- (.1,0.3) -- (-0.2,0.15) -- (.1,0);
            \node at (1,2) {$A$};
            \draw[very thick,shift={(2,2)},fill=red] (.1,-0.3) -- (.1,0.3) -- (-0.2,0.15) -- (.1,0);
            \node at (0,3) {2};
            \node at (1,3) {3};
            \node at (2,3) {2};
            \foreach \y in {1,2,3} {
                \node at (-1,\y) {1};
                \node at (3,\y) {1};
            };
        \end{tikzpicture}
        \caption{A consistent \ms\ instance. The square marked $A$ must contain a mine.}
        \label{fig:consistent}
    \end{subfigure}%
    \hspace{.04\linewidth}%
    \begin{subfigure}[t]{0.48\linewidth}
        \centering
        \begin{tikzpicture}[scale=0.5]
            \begin{scope}[shift={(-.5,-.5)}]
                \fill[gray!30] (0,2) rectangle (3,3);
                \draw[thick] (-1,0) grid (4,5);
            \end{scope}
            \node at (0,1) {2};
            \node at (1,1) {3};
            \node at (2,1) {2};
            \draw[very thick,shift={(0,2)},fill=red] (.1,-0.3) -- (.1,0.3) -- (-0.2,0.15) -- (.1,0);
            \node at (1,2) {$A$};
            \draw[very thick,shift={(2,2)},fill=red] (.1,-0.3) -- (.1,0.3) -- (-0.2,0.15) -- (.1,0);
            \node at (0,3) {1};
            \node at (1,3) {2};
            \node at (2,3) {1};
            \foreach \y in {1,2,3} {
                \node at (-1,\y) {1};
                \node at (3,\y) {1};
            };
        \end{tikzpicture}
        \caption{An inconsistent \ms\ instance. The uncovered sites above the flags imply $\neg A$, while those below the flags imply $A$.}
        \label{fig:inconsistent}
    \end{subfigure}%
    \caption{}
    \label{fig:consistency_example}
\end{figure}

In fact, \cons\ is NP-complete. This is shown by reduction from \sat, via the implementation of logic gates as \ms\ configurations which logically link the Boolean variables stored at the covered sites. For example, an AND gate would be constructed by setting up a \ms\ instance with two covered sites $a$ and $b$, and a third covered site $c$, with labels on the uncovered sites which together imply that $c$ contains a mine only if both $a$ and $b$ contain mines. Detailed constructions of such an AND gate as well as a NOT gate are known \cite{Kaye2000MinesweeperIN}, and with these components any Boolean circuit can be built. One can construct arbitrarily complex circuits on an infinite \ms\ board, and so \cons\ on an infinite board is Turing-complete \cite{Kaye2000MinesweeperTuring}.

As the authors of \cite{Scott2011MinesweeperMN} note, solving \cons\ is not directly relevant to a \ms\ player, who is promised a consistent configuration. Instead, the \ms\ player is tasked with deciding whether there exists a covered square which can be inferred to contain a mine, or to not contain a mine. We call this problem \infer. 

There is a simple reduction from \infer\ to the complement of \cons. Given a \ms\ instance, we iterate over all the covered sites and tentatively assume they are either empty or contain mines. For each of these tentative assignments, we consult a \cons\ oracle; if the oracle ever tells us we have an inconsistent board, then our tentative assignment is incorrect and we make the opposite inference. Since \cons\ is in NP, this reduction shows that \infer\ is in co-NP. Moreover, by reduction from \unsat, it has been shown that \infer\ is co-NP-complete \cite{Scott2011MinesweeperMN}.

\section{Phase Transition}\label{sec:transition}

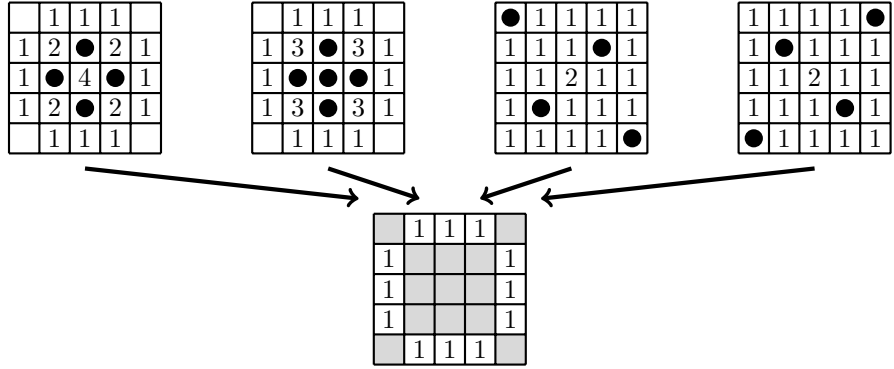
\begin{figure}
    \centering
    \begin{tikzpicture}
        \begin{scope}[scale=0.4]
            \begin{scope}[shift={(-.5,-.5)}]
                \fill[gray!30] (0,0) rectangle (1,1);
                \fill[gray!30] (5,0) rectangle (4,1);
                \fill[gray!30] (5,5) rectangle (4,4);
                \fill[gray!30] (0,5) rectangle (1,4);
                \fill[gray!30] (1,1) rectangle (4,4);
                \draw[thick] (0,0) grid (5,5);
            \end{scope}
            \node at (1,0) {1};
            \node at (2,0) {1};
            \node at (3,0) {1};
            \node at (1,4) {1};
            \node at (2,4) {1};
            \node at (3,4) {1};
            \node at (0,1) {1};
            \node at (0,2) {1};
            \node at (0,3) {1};
            \node at (4,1) {1};
            \node at (4,2) {1};
            \node at (4,3) {1};
            \draw[ultra thick,->] (-10,6) -- (-1,5);
            \draw[ultra thick,->] (-2,6) -- (1,5);
            \draw[ultra thick,->] (6,6) -- (3,5);
            \draw[ultra thick,->] (14,6) -- (5,5);
        \end{scope}
        \begin{scope}[scale=0.4,shift={(-12,7)}]
            \begin{scope}[shift={(-.5,-.5)}]
                \draw[thick] (0,0) grid (5,5);
            \end{scope}
            \node at (1,0) {1};
            \node at (2,0) {1};
            \node at (3,0) {1};
            \node at (1,4) {1};
            \node at (2,4) {1};
            \node at (3,4) {1};
            \node at (0,1) {1};
            \node at (0,2) {1};
            \node at (0,3) {1};
            \node at (4,1) {1};
            \node at (4,2) {1};
            \node at (4,3) {1};
            \node at (1,1) {2};
            \node at (3,1) {2};
            \node at (1,3) {2};
            \node at (3,3) {2};
            \node at (2,2) {4};
            \fill (2,1) circle (0.3);
            \fill (2,3) circle (0.3);
            \fill (1,2) circle (0.3);
            \fill (3,2) circle (0.3);
        \end{scope}
        \begin{scope}[scale=0.4,shift={(-4,7)}]
            \begin{scope}[shift={(-.5,-.5)}]
                \draw[thick] (0,0) grid (5,5);
            \end{scope}
            \node at (1,0) {1};
            \node at (2,0) {1};
            \node at (3,0) {1};
            \node at (1,4) {1};
            \node at (2,4) {1};
            \node at (3,4) {1};
            \node at (0,1) {1};
            \node at (0,2) {1};
            \node at (0,3) {1};
            \node at (4,1) {1};
            \node at (4,2) {1};
            \node at (4,3) {1};
            \node at (1,1) {3};
            \node at (3,1) {3};
            \node at (1,3) {3};
            \node at (3,3) {3};
            \fill (2,1) circle (0.3);
            \fill (2,3) circle (0.3);
            \fill (1,2) circle (0.3);
            \fill (3,2) circle (0.3);
            \fill (2,2) circle (0.3);
        \end{scope}
        \begin{scope}[scale=0.4,shift={(4,7)}]
            \begin{scope}[shift={(-.5,-.5)}]
                \draw[thick] (0,0) grid (5,5);
            \end{scope}
            \node at (1,0) {1};
            \node at (2,0) {1};
            \node at (3,0) {1};
            \node at (1,4) {1};
            \node at (2,4) {1};
            \node at (3,4) {1};
            \node at (0,1) {1};
            \node at (0,2) {1};
            \node at (0,3) {1};
            \node at (4,1) {1};
            \node at (4,2) {1};
            \node at (4,3) {1};
            \node at (2,1) {1};
            \node at (1,2) {1};
            \node at (3,2) {1};
            \node at (2,3) {1};
            \node at (1,3) {1};
            \node at (3,1) {1};
            \node at (0,0) {1};
            \node at (4,4) {1};
            \node at (2,2) {2};
            \fill (1,1) circle (0.3);
            \fill (3,3) circle (0.3);
            \fill (0,4) circle (0.3);
            \fill (4,0) circle (0.3);
        \end{scope}
        \begin{scope}[scale=0.4,shift={(12,7)}]
            \begin{scope}[shift={(-.5,-.5)}]
                \draw[thick] (0,0) grid (5,5);
            \end{scope}
            \node at (1,0) {1};
            \node at (2,0) {1};
            \node at (3,0) {1};
            \node at (1,4) {1};
            \node at (2,4) {1};
            \node at (3,4) {1};
            \node at (0,1) {1};
            \node at (0,2) {1};
            \node at (0,3) {1};
            \node at (4,1) {1};
            \node at (4,2) {1};
            \node at (4,3) {1};
            \node at (2,1) {1};
            \node at (1,2) {1};
            \node at (3,2) {1};
            \node at (2,3) {1};
            \node at (2,2) {2};
            \node at (1,1) {1};
            \node at (3,3) {1};
            \node at (0,4) {1};
            \node at (4,0) {1};
            \fill (1,3) circle (0.3);
            \fill (3,1) circle (0.3);
            \fill (0,0) circle (0.3);
            \fill (4,4) circle (0.3);
        \end{scope}
    \end{tikzpicture}
    \caption{No inference is possible in the \ms\ instance in the bottom panel, since the instance could have arisen from any of the boards shown above (among others), and for each covered site there is a consistent configuration in which it contains a mine and another in which it is empty. Note that we are not employing periodic boundary conditions here; alternatively, this grid should be thought of as part of a large \ms\ instance.}
    \label{fig:crapshoot}
\end{figure}

If $\text{P}\neq\text{NP}$, the results of Sec. \ref{sec:complexity} imply that no polynomial-time algorithm can decide \infer\ for all instances. Nonetheless, specific instances of \ms\ may yield to a polynomial-time approach, and experience suggests that this is the case for sufficiently low mine density. By contrast, at sufficiently high mine density, we expect to encounter configurations like that in Figure \ref{fig:crapshoot} where no inference is possible.

For intermediate densities, then, \ms\ must go from being easily solvable by inference to rarely solvable by inference. This claim is simple to test empirically. We start with \ms\ boards with a single site guaranteed to the player to be labeled zero, in order to remove uncertainty arising from the first move \cite{becerra2015algorithmic}. We use the reduction from \infer\ to \unsat\ outlined in the previous section to play \ms, deducing an empty or mined site at each step until either the game is won or no more inferences are possible. Our algorithm is outlined as follows.
\begin{enumerate}
    \item Start by uncovering the guaranteed 0-labeled site.
    \item Collect all covered squares bordering uncovered squares into a list \texttt{frontier\_outer}, and all uncovered squares bordering covered squares into a list \texttt{frontier\_inner}.
    \item Assign a Boolean variable to each site in \texttt{frontier\_outer}.
    \item From each site $i$ in \texttt{frontier\_inner}, construct a Boolean constraint $F_i$, and transform it into conjunctive normal form (CNF). For example, if a site has effective label 1 and borders two covered sites assigned to variables $x_\beta$ and $x_\gamma$ in the previous step, then we have
    \begin{equation}
        (\neg x_\beta \wedge x_\gamma)\vee (x_\beta\wedge \neg x_\gamma),
    \end{equation}
    which can be written in CNF as
    \begin{equation}
        F_i = (x_\beta\vee x_\gamma)\wedge (\neg x_\beta\vee \neg x_\gamma).
    \end{equation}
    \item Let $F = \bigwedge_{i\in I} F_i$, with $I$ an index set for \texttt{frontier\_inner}, be the conjunction of all the formulas constructed in the previous step. For every site in \texttt{frontier\_outer}, with Boolean variable $x_\beta$, test $F\wedge x_\beta$ for satisfiability. If it is unsatisfiable, then this site must not have a mine, so reveal it. If it is satisfiable, then test $F\wedge \neg x_\beta$ for satisfiability. If it is unsatisfiable, then this site must have a mine, so flag it.
    \item If all mines are flagged, or if no inferences are found, conclude. Otherwise, return to step 2.
\end{enumerate}

\begin{figure}[t]
    \centering
    \includegraphics[width=0.9\linewidth]{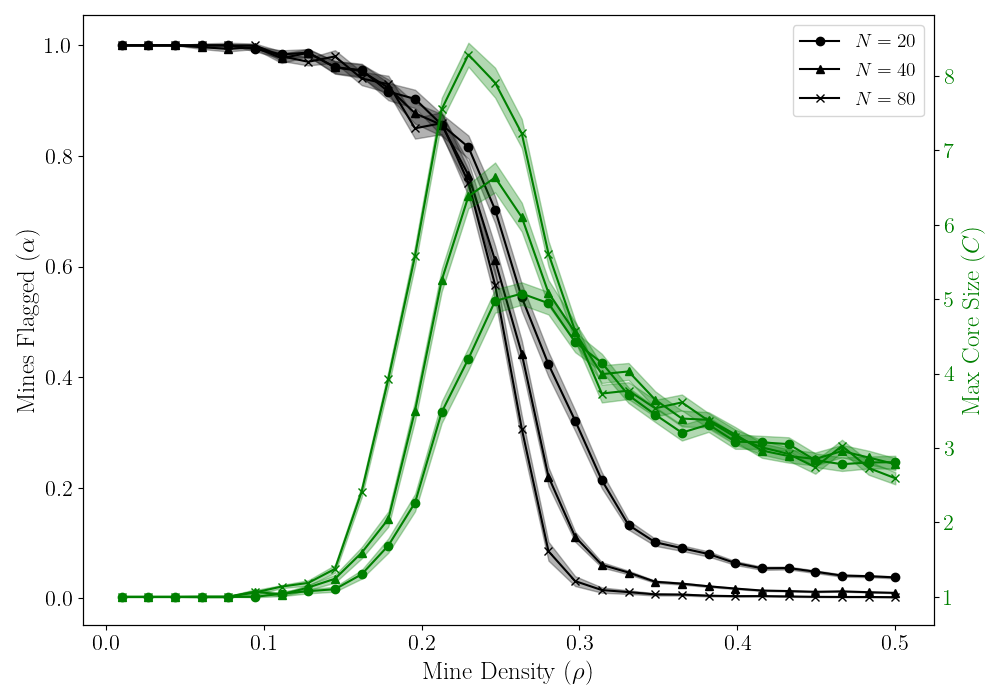}
    \caption{Averaged results and standard errors of 300 \ms\ games at each mine density for each grid size $N = 20$, 40 and 80 are shown. In black, we plot the expected fraction $\alpha$ of mines flagged by the \infer\ solver as a function of the mine density $\rho$. In green, we plot the expected maximum size of the grouped minimal unsatisfiable (GMUS) cores encountered in the reduction to \sat. This is a measure of the hardness of \ms\ as a function of $\rho$, and roughly indicates the number of labeled sites one needs to look at in order to make inferences.}
    \label{fig:data}
\end{figure}

In Figure \ref{fig:data}, we show the results of these tests. For grids of sizes $N = 20$, 40, and 80, we plot the empirical fraction $\alpha$ of mines flagged by inference. We find that $\alpha$ rapidly declines to 0 roughly in the interval $0.2<\rho<0.3$, and the decline is steeper at higher values of $N$.

We would also like to determine how difficult it is to solve the \ms\ inference problem at various mine densities. We use an algorithm-agnostic metric derived from the reduction to \unsat. As described in the steps above, we construct a Boolean formula $F_i$ from every labeled site in the inner frontier. We then test the conjunction of all these formulas, together with one tentative assumption:
\begin{equation}
    \bigwedge_{i\in I} F_i \wedge x_\beta, \qquad\text{or}\qquad \bigwedge_{i\in I} F_i \wedge \neg x_\beta,
\end{equation}
where $\beta$ is the index of a covered site on the outer frontier and $I$ indexes the inner frontier. If one of these formulas is unsatisfiable, then we have proved by contradiction the presence or absence of a mine. In this case we can extract the \emph{grouped minimal unsatisfiable (GMUS) core}, a subset $S\in I$ for which $\bigwedge_{i\in S} F_i \wedge x_\beta$ is unsatisfiable, but for every proper subset $S'\subsetneq S$ the formula $\bigwedge_{i\in S'} F_i \wedge x_\beta$ is satisfiable (or the equivalent with $\neg x_\beta$ in place of $x_\beta$). As a measure of hardness we use $C = |S|$, the number of labeled sites one needs to look at in order to make the inference about the presence or absence of a mine at site $\beta$.

Figure \ref{fig:data} shows the average value of the maximum GMUS core size encountered during the \ms\ games as a function of the mine density. This is computed using the open-source \texttt{MUSer2} library \cite{Belov2012MUSer2AE}. The core size is strongly peaked in the vicinity of the phase transition, and the peak core size grows with the size of the lattice. This indicates that playing \ms\ near the critical mine density requires looking at large patches of the board to make inferences. Furthermore, we clearly see the easy-hard-easy pattern first suggested in \cite{cheeseman1991really}.

\section{Algorithmic Barriers}\label{sec:algorithms}

In Section \ref{sec:transition} we showed that the maximum GMUS core size encountered during the course of a \ms\ game peaks at the phase transition. This suggests that the hardest \ms\ instances arise at the critical mine density. As another test of this hypothesis, we present a family of polynomial-time approaches to \infer. These algorithms partially solve \infer, in the sense that they can identify some but not all inferences, and never make an invalid inference; there are false negatives, but no false positives. We show that they perform well below the critical mine density, but fail to find a significant proportion of inferences at or near the critical density.

Our algorithms are a generalization of the ``na\"ive single-point'' algorithm in \cite{becerra2015algorithmic}. In the na\"ive single-point approach, only two types of inferences can be made:
\begin{itemize}
    \item If the effective label of a site is 0, then any of its covered neighbors can be inferred to be safe.
    \item If the effective label of a site is equal to its number of covered neighbors, then any of its covered neighbors can be inferred to be mines.
\end{itemize}
These sorts of inferences are possible precisely when $C = 1$, i.e., when the \sat\ solver need only use a single labeled site to make an inference.

\begin{figure}[t]
    \centering
    \includegraphics[width=0.9\linewidth]{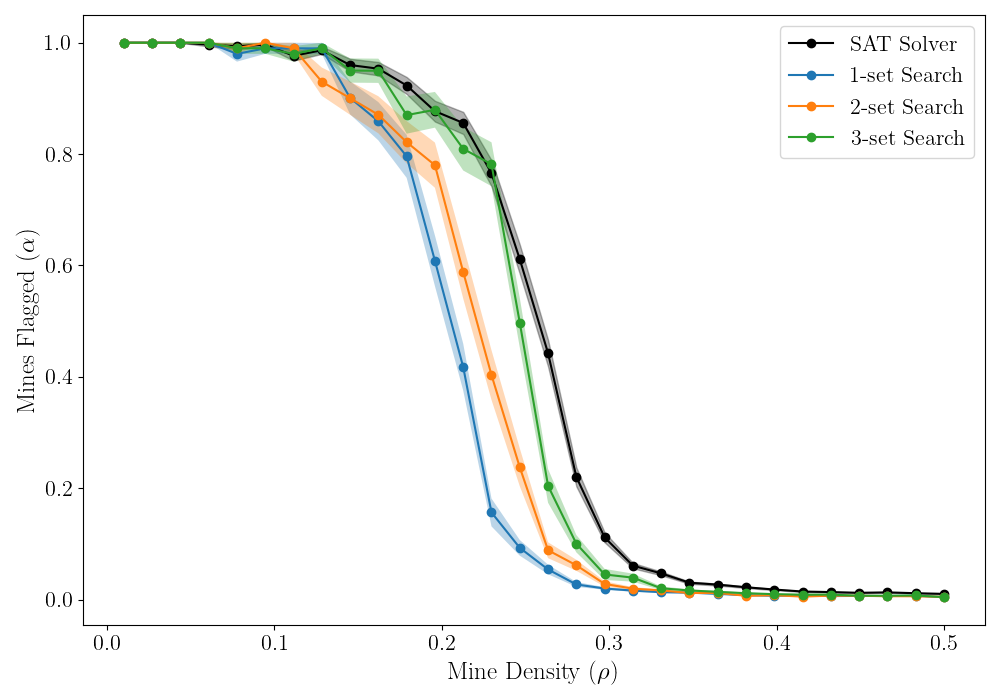}
    \caption{The performance of the \sat\ solver is compared to the polynomial-time $k$-set search algorithm on a board of size $N=40$. The \sat\ solver finds the theoretical maximum fraction of mines which can be flagged by inference. The $k$-set search approach performs comparably until the phase transition at $0.2<\rho<0.3$, at which point the \sat\ solver clearly outperforms $k$-set search, and moreover the polynomial-time algorithms are stratified by $k$.}
    \label{fig:barriers}
\end{figure}

The na\"ive single-point algorithm has $\mathcal{O}\left(n_\text{inner}\right)$ complexity on a single \ms\ instance, where $n_\text{inner}$ is the size of the inner frontier. We employ a natural generalization with $\mathcal{O}\left(n_\text{inner}^k\right)$ complexity, which we call $k$-set search. We label the sites of the outer frontier by $j = 1,\ldots,n_\text{outer}$, and assign to each site $i=1,\ldots,n_\text{inner}$ in the inner frontier a vector
\begin{equation}\label{eq:hg_incidence}
    a_{ij} = \begin{cases} 1 & \text{if $i$ borders $j$} \\ 0 & \text{otherwise} \end{cases},
\end{equation}
and denote the effective label of site $i$ with $e_i$. In this notation, the constraint provided by a site in the inner frontier is
\begin{equation}
    \sum_{j=1}^{n_\text{outer}} a_{ij}x_j = e_i.
\end{equation}
We then iterate over all combinations of up to $k$ of these vectors, with replacement, denoted by index sets $\{i_1,\ldots,i_k\}$. For each such combination, we consider all the linear combinations of constraints,
\begin{equation}
    \sum_{\ell=1}^k \left(-1\right)^{b_\ell} \sum_{j=1}^{n_\text{outer}} a_{i_\ell j} x_j = \sum_{\ell=1}^k \left(-1\right)^{b_\ell} e_{i_\ell},
\end{equation}
with $b_\ell = 0,1$. If the right hand side is equal to either the minimum or maximum value of the left hand side, then we can infer the value of any of the variables appearing on the left hand side with a nonzero coefficient.

It is simple to see that any inference made by $k$-set search could also be made by a \sat\ solver limited to a GMUS core size $C \le k$. The converse is almost true: if the GMUS core has size $k$, $k$-set search will only fail if the inference comes from a linear combination of constraints with coefficients other than $\pm 1$, which is rare. We thus expect that $k$-set search will perform well up until the point where the core size reaches $k$. Figure \ref{fig:data} shows that the core size is peaked around the phase transition, so in practice we expect that $k$-set search will perform significantly less well than a full \sat\ solver in the vicinity of the phase transition.

Figure \ref{fig:barriers} confirms this claim. We plot the average fraction $\alpha$ of mines flagged by 1-, 2-, and 3-set search, compared to the performance of the \sat\ solver. As we expect, all algorithms perform roughly equally at low $\rho$, and then stratify by $k$ in the vicinity of the phase transition. In the interval $0.2<\rho<0.3$, a \ms\ player would benefit substantially from using $(k+1)$-set search over $k$-set search.

\section{Asymptotic Behavior}\label{sec:asymptotic}

Naturally, one of the key questions regarding the \ms\ phase transition is its asymptotic behavior. Our empirical studies strongly suggest some kind of phase transition, loosely speaking, as $N\to\infty$. The question is whether this transition is sharp. Let $\alpha(N, \rho)$ denote the expected fraction of mines which can be flagged by inference on a board of size $N$ with mine density $\rho$, as in Figure \ref{fig:data}. Following \cite{friedgut1999sharp}, we say the \ms\ transition is coarse if there exists some constant $C>0$ such that, for all $\rho$,
\begin{equation}
    \lim_{N\to\infty}\left| \rho \d{\alpha(N, \rho)}{\rho}\right| \le C.
\end{equation}
Otherwise, we say the transition is sharp.

For several reasons, it is difficult to directly address the question of whether \ms\ has a sharp phase transition. First, the distribution of \ms\ instances is quite complicated. Although it is simple to define a uniformly random distribution of \ms\ boards, the \ms\ instances which arise during game-play depend on a player's strategy. 

\begin{figure}
    \centering
    \includegraphics[width=\linewidth]{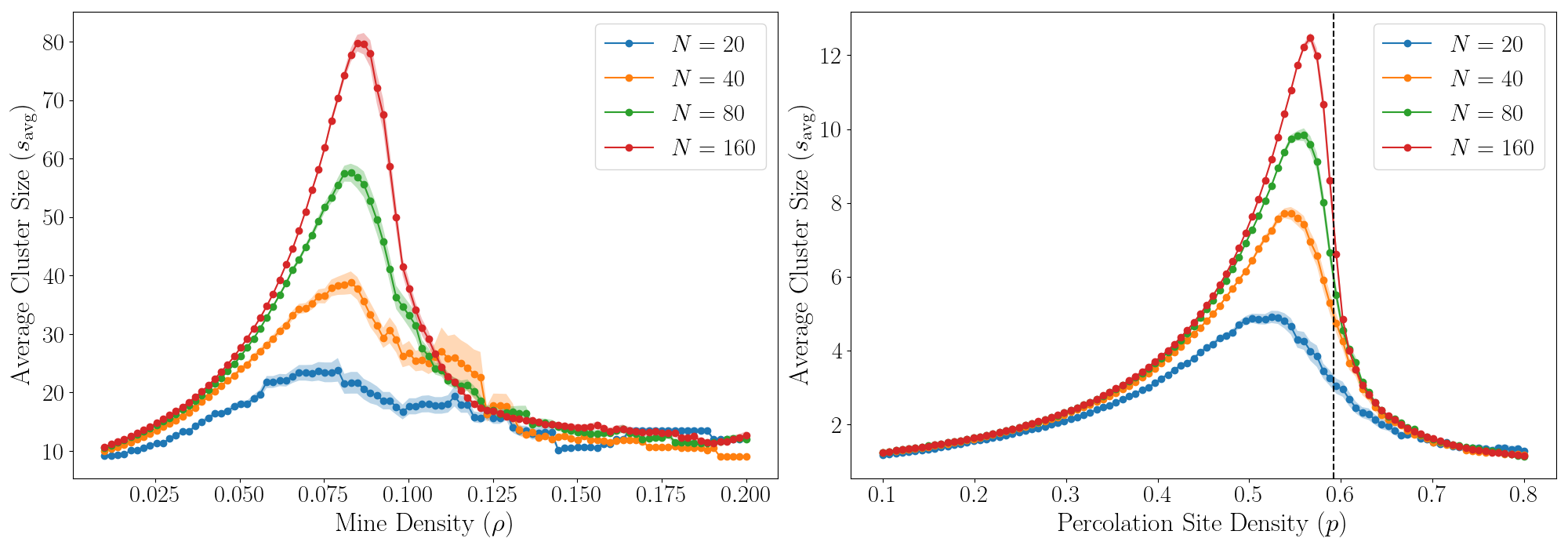}
    \caption{The average cluster sizes appearing in the standard percolation problem (right) for a range of values of the site probability, and in the sites with mines or nonzero labels in \ms\ (left) for a range of values of the mine density. The vertical line on the right shows the threshold probabiilty for percolation.}
    \label{fig:correlation}
\end{figure}

Even if we restrict to the case of the perfectly logical player who operates by solving \unsat, the frontiers which arise depend on the geometry of the percolation clusters formed by the mines and the sites with nonzero labels. The probability of a given site having either a mine or a nonzero label is
\begin{equation}
    P\left(\text{mine or nonzero}\right) = 1-P(\text{zero}) = 1-(1-\rho)^9.
\end{equation}
Setting this equal to the threshold value of $p \approx 0.59$ for site percolation on a 2D square lattice \cite{stauffer1994introduction}, we find $\rho \approx 0.1$. However, we are not justified in making such a direct comparison to the standard percolation problem, because nearby sites in \ms\ are correlated. To address this, we perform Monte Carlo simulations of the formation of percolation clusters in \ms. Figure \ref{fig:correlation} shows the average cluster size $s_\text{avg}$, defined as in \cite{christensen2002percolation}, of the clusters which form on \ms\ boards compared to those in standard percolation theory. The behavior at $\rho \approx 0.1$ shows that our calculation gives roughly the correct location of the percolation threshold for \ms.

The geometry of the percolation clusters in the vicinity of the \ms\ percolation threshold is related to that of the clusters in standard percolation theory with independent sites. This follows from the general principle of universality: the ``microscopic'' details of a system do not affect bulk properties at the critical point. More precisely, renormalization group methods show that if correlations between sites fall off faster than $r^{-d}$, then the correlations are irrelevant at the critical point as long as $d\nu > 2$, where $\nu$ is the critical exponent for correlation length \cite{Saberi_2015}. The correlations in \ms\ have finite range, so they decay faster than any power law, and hence we can ignore the correlations when considering the geometry of the frontier near the percolation threshold. It is well-known that percolation clusters exhibit complex fractal geometry near the percolation threshold \cite{Strelniker2009}, and so this complexity should carry over to random \ms\ instances.

All this is simply to argue that random \ms\ instances sample in a very complicated way from the space of possible \infer\ problems. There are thus two questions of separate import. First we should ask if \infer\ itself, or some generalization of it, has a sharp phase transition with respect to some parametrized distribution of instances. If this is answered in the affirmative, then the next question is whether \ms\ itself samples from the space of \infer\ problems in such a way as to lead to a phase transition with respect to the mine density $\rho$.

It seems likely that \infer\ itself exhibits a phase transition. To see this, we consider a space of problems which contains \infer\ as a special case, and for which it is much simpler to define  random instances. We start by noting that \eqref{eq:hg_incidence} defines the incidence matrix of a hypergraph where vertices correspond to covered sites in the outer frontier and hyperedges correspond to labeled sites in the inner frontier. Each hyperedge can be assigned the effective label of its corresponding site, and then \infer\ becomes the following question: if each vertex must be assigned a value 0 or 1, such that the sum of vertices in each hyperedge is equal to its label, is the value of any vertex constrained to be either 0 or 1?

The natural generalization is to replace the hypergraph defined by \eqref{eq:hg_incidence} with an arbitrary hypergraph, and the effective labels with an integer label between 0 and $|e|$ on each hyperedge $e$. We may define a probability measure $\mu$ on hypergraphs for which hyperedges of size $m$ appear independently with probability $p_m$.

The presence of a sharp phase transition in properties of graphs or hypergraphs can be shown using Friedgut's theorem \cite{friedgut2005hunting}. Roughly speaking, this theorem establishes that any property of hypergraphs which does not exhibit a sharp phase transition must be approximable by a local property. We will not prove the existence of a sharp phase transition for inference on random hypergraphs, but intuitively, it is clear that the inference property should not be locally approximable. Inferences can arise from large sets of hyperedges in myriad ways; we cannot expect to account for almost all inferences by checking for a finite set of sub-hypergraphs. Indeed, the $k$-set search algorithm of Section \ref{sec:algorithms} effectively checks for a finite set of sub-hypergraphs, and for any fixed $k$ we do not expect this algorithm to perform well for sufficiently large $N$. These considerations suggest that inference on random hypergraphs has a sharp phase transition at some hyperedge probability $p_*(N)$.

Returning to the question of the \ms\ phase transition, some simple estimates suggest that even if inference on random hypergraphs does have a sharp phase transition, the behavior of the \ms\ phase transition depends on details which are difficult to ascertain. The distribution of hypergraphs which appear in the inference problems in \ms\ is certainly a complicated function of the mine density, and even for fixed mine density, overlapping hyperedges are not independent of one another. Thus, throughout a game of \ms, the \infer\ problems encountered by a player are sampled from distributions of hypergraph inference problems with a range of parameters.

As a simple phenomenological model of this behavior, assume that the \infer\ problems which appear during a game of \ms\ are sampled using some parameter $p$, and that there is a sharp threshold at $p = p_*(N)$ above which inferences become possible. Furthermore, assume the parameter $p$ is itself drawn from some distribution $f(p; N,\rho)$. The game of \ms\ will continue until the random value of $p$ falls below $p_*(N)$. On each turn, this happens with probability
\begin{equation}
    P(N,\rho) = \int_0^{p_*(N)} f(p; N,\rho)\,dp.
\end{equation}
The expected fraction of mines flagged then scales as the number of turns played, so
\begin{equation}\label{eq:model}
    \alpha(N, \rho) \sim \frac{1}{N}\sum_{n=0}^{T(N,\rho)} n (1-P(N,\rho))^n P(N,\rho) \sim (1-P(N, \rho))^{T(N,\rho)},
\end{equation}
where $T(N,\rho)$ is the expected number of turns required to complete the game if all mines are flagged. As $\rho$ is increased, we expect $P$ to increase, since under-constrained inference problems like the one in Figure \ref{fig:crapshoot} are more likely to appear. Thus, \eqref{eq:model} reproduces the sigmoidal behavior evident in Figure \ref{fig:data}. 

As $N$ is taken to infinity, the shape of $\alpha(N,\rho)$ depends on the exact limiting behavior of $P(N, \rho)$ and $T(N,\rho)$, which in turn depend on the detailed statistics of \ms\ instances. It may approach either a discontinuous function, corresponding to a sharp phase transition, or some curve which interpolates between $\alpha = 1$ and $\alpha = 0$ over a finite range of mine densities, giving a coarse phase transition.

Despite this uncertainty in the asymptotic behavior, our empirical results show that the phase transition in \ms\ at finite $N$ is closely associated with the appearance of hard \infer\ problems. We can thus understand a great deal about the average-case complexity of \ms\ by thinking in terms of its phase transition.

\bibliography{main}

\end{document}